\documentclass[]{fairmeta}
\usepackage{xspace}
\usepackage{amsmath}
\usepackage{amssymb}
\usepackage{booktabs}
\usepackage{adjustbox}
\usepackage{threeparttable}
\usepackage{tabularx}
\usepackage{xcolor}
\usepackage{xurl}
\usepackage{comment}
\newcommand{\imp}[1]{#1\,\textcolor{green!55!black}{\scriptsize$\uparrow$}}

\newcommand{\reg}[1]{#1\,\textcolor{red!75!black}{\scriptsize$\downarrow$}}
\title{Remember When It Matters: Proactive Memory Agent for Long-Horizon Agents}

\author{Yifan Wu}
\author{Lizhu Zhang}
\author{Yuhang Zhou}
\author{Mingyi Wang}
\author{Bo Peng}
\author{Serena Li}
\author{Xiangjun Fan}
\author{Zhuokai Zhao}
\affiliation{Meta AI}

\abstract{In long-horizon tasks, decision-relevant state is often scattered across an expanding trajectory, while the action agent must surface it and act.
As trajectories grow, task requirements, environment facts, prior attempts, diagnoses, and open subgoals can be buried in the context window or pushed beyond it, failing to influence decisions when needed.
We call this failure mode \emph{behavioral state decay}.
We study memory as an active intervention mechanism rather than passive retrieval.
A separate memory agent runs alongside an unmodified action agent, updating a structured memory bank from the recent trajectory and deciding whether to inject a memory-grounded reminder or remain silent.
The module is plug-and-play with frontier action agents and existing agent harnesses.
Across Terminal-Bench 2.0 and $\tau^2$-Bench, it improves pass@1 for both weaker and stronger action agents, with gains of +8.3~pp on Terminal-Bench and +6.8~pp on $\tau^2$-Bench.
Ablations show that selective intervention outperforms passive bank exposure, always-on injection, advisor-only guidance, and general retrieval.
As an early step toward open-weight memory policies, we train Qwen3.5-27B on SETA using SFT and GRPO, improving validation reward and achieving partial transfer to Terminal-Bench.}

\date{\today}
\correspondence{Yifan Wu and Zhuokai Zhao at \email{\{yfwu, zhuokai\}@meta.com}}

\metadata[Code]{\url{https://github.com/yifannnwu/proactive-memory-agent}}

\begin{document}

\maketitle

\section{Introduction}
\label{section:intro}

LLM agents are increasingly evaluated on long-horizon tasks that require many rounds of tool use and environment interaction~\citep{yao2023react,shinn2023reflexion,wang2024voyager}. These tasks span autonomous command-line execution~\citep{merrill2026terminalbench}, multi-step machine-learning engineering~\citep{chan2024mlebench}, and interactive tool use under domain-specific rules~\citep{yao2024taubench,barres2025tau2bench}. Because such tasks unfold over many observations, actions, and partial decisions, success depends not only on solving each local problem, but also on preserving information that should constrain future behavior. Yet agents often fail in ways that reveal a breakdown in this state maintenance: they may identify a requirement early in a task and later violate it while fixing an unrelated bug; observe that a command, parameter setting, or implementation path fails and later retry a near-identical variant; or diagnose an error pattern and later treat the same pattern as new. These failures suggest that simply making longer histories available is insufficient. Long-horizon agents need mechanisms for keeping decision-relevant execution state active over time.

We call this failure mode \textbf{behavioral state decay}: during long-horizon execution, information that should shape future actions like task requirements, environment facts, previous attempts, failure diagnoses, intermediate discoveries, and open subgoals stops influencing the agent's next decision. The information may still be present in the transcript, or may even remain within the model's context window, but it no longer exerts reliable control over behavior \citep{liu2024lost}. This distinction is important because many memory approaches focus on whether information can be stored or retrieved, while long-horizon task execution also requires deciding when remembered information should affect the agent's next action.

A natural response is to equip agents with memory. Existing memory systems typically emphasize storing, updating, and retrieving relevant records \citep{packer2023memgpt,zhong2024memorybank,park2023generative,mem0ai2026mem0}, which is crucial for personalization, persistent user state, and cross-session recall. However, active task execution introduces a different problem: memory must decide \emph{when} to intervene. Surfacing too little memory lets the agent repeat mistakes or ignore prior discoveries; surfacing too much adds latency, consumes tokens, and can distract the agent from local progress. Effective memory for long-horizon agents is not only a write and retrieval problem, but also an intervention problem.

This is a stronger control question than summarization. A summarizer asks what to retain; our memory asks whether any retained execution state should become active in the action agent's next decision. Long-horizon tasks vary in their failure modes---in some, the useful memory is a hard requirement from the initial instruction; in others, an environment fact, a failed command, a bug diagnosis, or an unfinished subgoal---so a fixed summarization policy cannot know whether a memory should interrupt the next action.

We study memory as intervention. We introduce a memory agent that runs alongside an unmodified action agent. At fixed intervals, it updates a structured memory bank from the recent trajectory and then decides whether to inject a concise reminder---a forgotten requirement, a stable environment fact, a failed attempt, or a diagnosis---into the next action-agent call, or to remain silent. Unlike passive memory systems, it does not just retrieve records on demand; it decides whether memory should enter the control loop \citep{zhang2025memact}. Unlike general advisor models \citep{asawa2025advisor,anthropic2026advisor}, it is constrained to memory-grounded reminders rather than broad strategic guidance.

We evaluate this architecture on Terminal-Bench 2.0 \citep{merrill2026terminalbench} and $\tau^2$-Bench \citep{yao2024taubench,barres2025tau2bench}, covering both autonomous command-line execution and interactive tool-use. With Claude Opus 4.6 as the memory agent, memory intervention improves Claude Sonnet 4.5 on Terminal-Bench (37.6\%~$\to$~45.9\%) and on $\tau^2$-Bench (55.0\%~$\to$~61.8\%), and the gain does not vanish for a stronger action agent: Opus 4.6 still gains +2.4~pp and +2.5~pp respectively. Ablations show that our proactive intervention beats passive memory exposure and general memory retrieval. As an early step toward an open-weight memory policy, we further fine-tune Qwen3.5-27B as the memory agent with SFT and GRPO and observe that gains transfer to held-out Terminal-Bench.

Our contributions are: (i) we identify \emph{behavioral state decay} as a central failure mode of long-horizon language agents; (ii) we propose a two-phase memory intervention architecture that decouples memory maintenance from action selection. We show empirically that memory intervention improves long-horizon task performance and that ablations favor maintained memory plus selective intervention; (iii) as preliminary evidence, we show that the intervention policy can be partially learned by an open-weight model.

\section{Related Work}

\subsection{Long-horizon language agents.}
Language-model agents are commonly instantiated as interleaved reasoning-and-acting systems that issue tool calls, observe environment feedback, and revise their plans over multiple steps \citep{yao2023react}. Recent benchmarks have pushed this setting toward realistic long-horizon tasks, including command-line environments \citep{merrill2026terminalbench}, tool-agent-user interactions with domain rules \citep{yao2024taubench}, machine-learning engineering \citep{chan2024mlebench}, etc. These settings expose failures that are not captured by single-turn reasoning benchmarks: agents may repeat failed actions, lose track of requirements, or drift away from earlier discoveries. Our work focuses on this form of behavioral state decay and studies memory as a mechanism for maintaining decision-relevant execution state during such trajectories.

\subsection{Memory for language-model agents.}
A large body of work augments language models with external or long-term memory. Retrieval-augmented models use non-parametric stores to improve factual grounding and knowledge access \citep{guu2020realm,lewis2020rag}, while long-context memory architectures cache or retrieve past representations to extend the effective context available to the model \citep{wang2023longmem}. Agent memory systems further organize past experience into natural-language records for long-term interaction, personalization, and behavior generation \citep{park2023generative,zhong2024memorybank,packer2023memgpt,zhang2024memorysurvey}. Lifelong agents such as Voyager store reusable skills and retrieve them for new tasks \citep{wang2024voyager}, and production memory layers such as Mem0 emphasize persistent user/session/agent memory with efficient storage and retrieval \citep{mem0ai2026mem0}. Our work is complementary to these systems. Rather than optimizing persistent memory storage or retrieval alone, we ask when remembered execution state should be reactivated as an intervention in an ongoing loop.

\subsection{Learned memory policies and context management.}
Long context windows do not by themselves guarantee reliable use of prior information: models can under-utilize information depending on its position in the context \citep{liu2024lost}. This has motivated work on context compression, memory editing, and learnable context curation. Memory-as-Action formulates working-memory management as explicit editing actions optimized jointly with task performance \citep{zhang2025memact}. Context-Folding lets an agent branch into sub-trajectories for subtasks and then fold completed branches into concise summaries, with reinforcement learning rewards for decomposition and context management \citep{sun2025contextfolding}. Mem-$\alpha$ trains agents to manage memory through tool-based operations over sequential information chunks, using downstream question-answering accuracy over the full interaction history as the reward signal \citep{wang2025memalpha}. Our work shares with these methods the view that memory and context management should be optimized for downstream outcomes, but targets a different control problem. We do not train the action agent to curate its own context, optimize memory construction for later question answering, or fold completed sub-tasks into summaries. Instead, a separate memory agent repeatedly observes a live multi-turn task trajectory, maintains structured execution state, and decides whether any part of that state should intervene in the next action-agent call. This creates a more heterogeneous credit-assignment problem: across tasks, memory may need to remain silent, prevent repeated failures, or reactivate diagnoses and open sub-goals, and unnecessary interventions can be harmful rather than merely redundant.

\subsection{Reflection, critics, and advisor models.}
Another line of work improves agents through feedback, reflection, or auxiliary reasoning. Self-Refine iteratively improves model outputs using self-generated feedback \citep{madaan2023selfrefine}; Reflexion stores verbal reflections from prior attempts in an episodic memory buffer \citep{shinn2023reflexion}; and Tree of Thoughts expands inference-time reasoning through search over intermediate thoughts \citep{yao2023tree}. More recently, advisor-style methods use a second model to steer a black-box or executor model with natural-language guidance \citep{asawa2025advisor,anthropic2026advisor}. Our memory agent is related to these supervisory approaches, but is more constrained: it is not intended to provide broad strategic advice or perform general deliberation. Its output is a memory-grounded reminder derived from maintained execution state, such as a task requirement, environment fact, failed attempt, diagnosis, or open subgoal that is at risk of becoming behaviorally inactive.

\subsection{Positioning.}
Taken together, prior work shows that memory, retrieval, reflection, context compression, and auxiliary guidance can improve language agents. Our contribution is to formulate memory as a selective intervention policy for long-horizon execution. The central question is not only what should be remembered or summarized, but whether, when, and how remembered execution state should enter the action agent's context so that it changes the next decision without unnecessary token or latency overhead. Because tasks differ substantially in environment dynamics, observability, and failure modes, this requires intervention calibration rather than a fixed summarization policy.

\section{Method}

\begin{figure*}[t]
\centering
\begin{subfigure}[t]{0.48\textwidth}
\centering
\includegraphics[width=\linewidth]{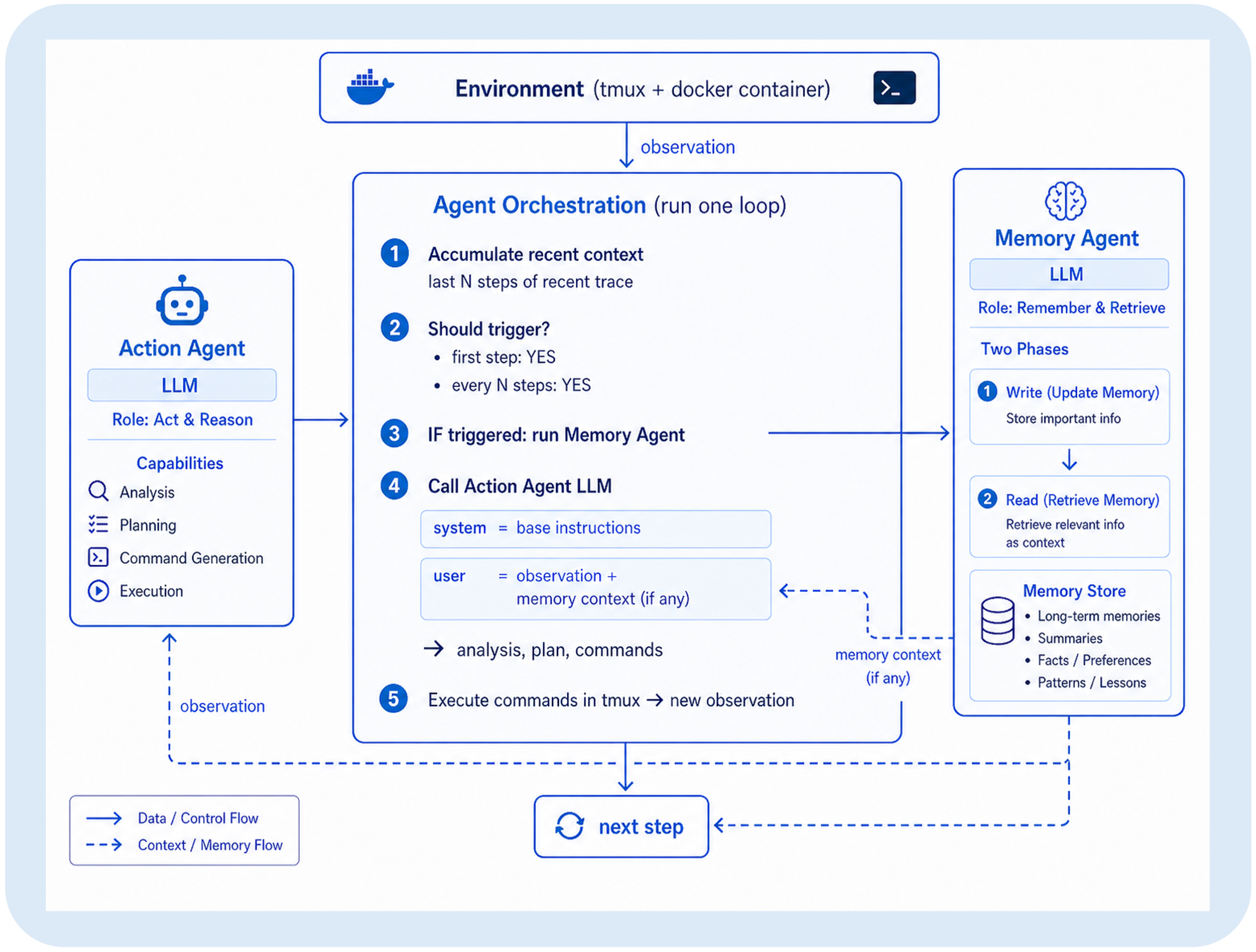}
\caption{System integration. The action agent (left) interacts with the
environment; the memory agent (right) runs alongside, observing a sliding
window of recent steps and the current memory store. At every $N$ steps the
memory agent is invoked to update the bank and optionally inject a context
reminder into the next action-agent call.}
\label{fig:method_system}
\end{subfigure}\hfill
\begin{subfigure}[t]{0.48\textwidth}
\centering
\includegraphics[width=\linewidth]{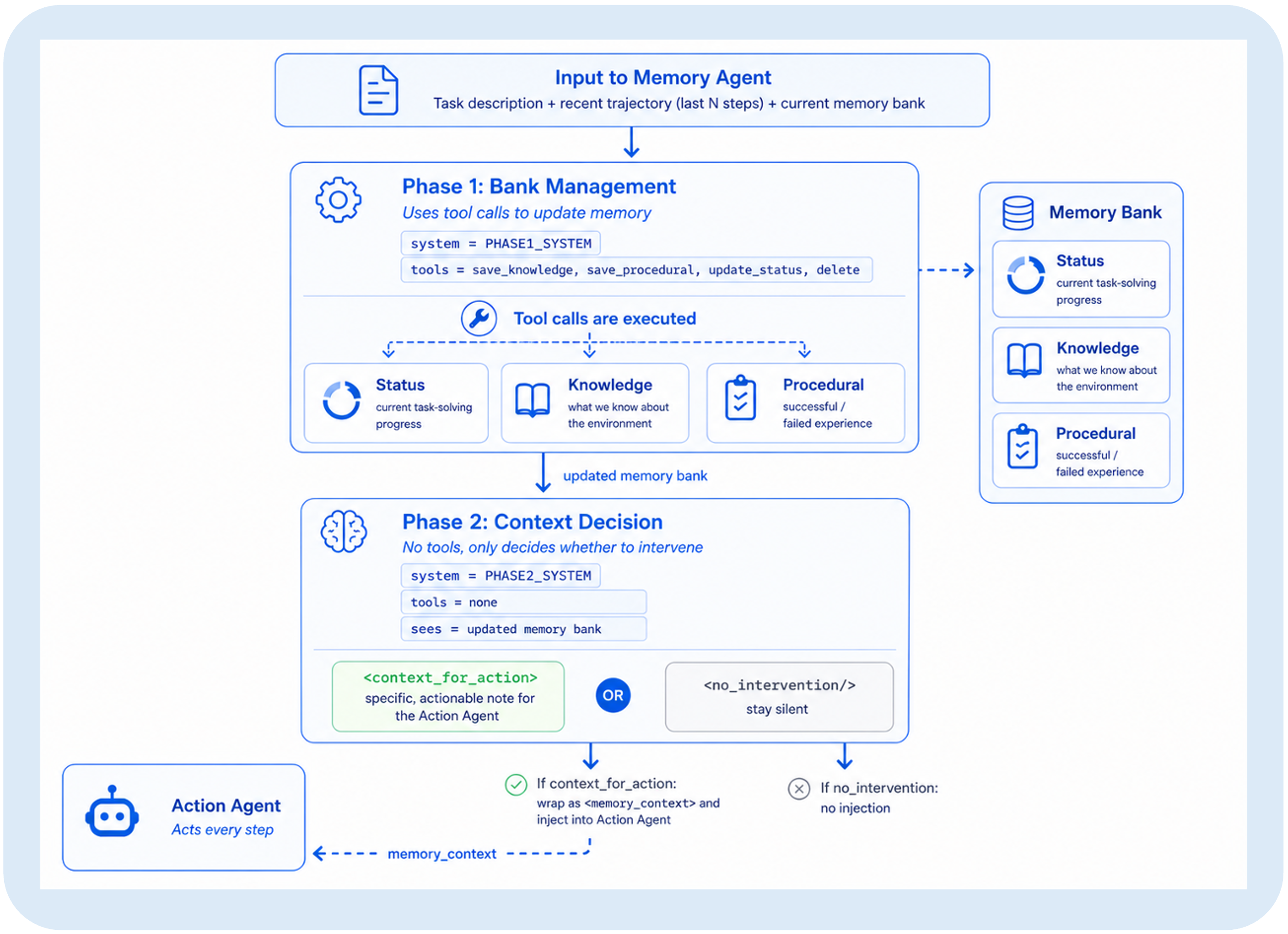}
\caption{Memory agent internals. Phase~1 manages the memory bank
(\emph{status}, \emph{knowledge}, \emph{procedural} entries) through tool
calls. Phase~2 reads the updated bank and either emits a
\texttt{<context\_for\_action>} reminder or \texttt{<no\_intervention/>}.}
\label{fig:method_memory_agent}
\end{subfigure}
\caption{Overview of the memory-intervention architecture. (a) The memory agent
runs as a separate process beside an unmodified action agent. (b) Within each
memory step, a two-phase workflow first updates the structured memory bank,
then decides whether any remembered state should enter the action agent's next
decision.}
\label{fig:method_overview}
\end{figure*}

We introduce a memory intervention architecture for long-horizon language agents.
The goal is to maintain decision-relevant execution state over a multi-step
trajectory and selectively reactivate that state when it is likely to affect the
action agent's next decision. Our method leaves the action agent unchanged and
adds a separate memory agent that observes the trajectory, updates a structured
memory bank, and decides whether to inject a concise reminder into the action
agent's context. Figure~\ref{fig:method_overview} summarizes the architecture:
the left panel shows how the memory agent is integrated with the action agent
in the run loop, and the right panel zooms into the memory agent's two-phase
internal workflow.

\subsection{Problem Setup}

We consider an action agent interacting with an environment over a trajectory
$\tau = (o_1, a_1, o_2, a_2, \ldots, o_T)$, where $o_t$ is the environment
observation at step $t$ and $a_t$ is the action produced by the agent. The
agent is given a task description $x$ and samples
$a_t \sim \pi_A(a_t \mid x, \tau_{<t})$, where $\pi_A$ is the action policy
implemented by an LLM and tool-use scaffold. In long-horizon settings, the
context the scaffold actually exposes to $\pi_A$ may be truncated, summarized,
or otherwise filtered, so not every piece of trajectory information remains
available at each step.

In long-horizon tasks, the full trajectory may contain many pieces of information
that should continue to influence future actions even when they are no longer
salient in the action-agent context: task requirements, environment facts, previous attempts,
failure diagnoses, intermediate discoveries, and open subgoals. We refer to this
information as \emph{execution state}. Our objective is to maintain execution
state separately from the action context and decide when it should be reintroduced
into the action loop.

We augment the action agent with a memory agent $\pi_M$. The memory agent is
invoked at the first step and then at a fixed interval. At each memory timestep
$t$, it observes the task description, a recent trajectory window
$w_t = \mathcal{W}_k(\tau_{<t}, o_t)$, and the current memory bank $B_{t-1}$.
It first updates the bank and then decides whether to emit an intervention:
\[
\begin{aligned}
B_t &\sim \pi_M^{\mathrm{edit}}(\cdot \mid x, w_t, B_{t-1}), \\
i_t &\sim \pi_M^{\mathrm{intervene}}(\cdot \mid x, w_t, B_t),
\end{aligned}
\]
where $i_t \in \{\varnothing, \text{text reminder}\}$. A non-null intervention
is injected into the next action-agent call as transient memory context;
otherwise the action agent proceeds without additional memory context.

This formulation treats memory as a policy over interventions. The memory agent
must decide not only what execution state to retain, but also when remembered
state is useful enough to enter the action loop.

\subsection{Memory Bank}

The memory bank is a compact, structured representation of execution state. It
contains three components:
\[
B_t = (s_t, K_t, P_t),
\]
where $s_t$ is a private status field, $K_t$ is a set of knowledge memories, and
$P_t$ is a set of procedural memories.

The \textbf{status} field tracks the memory agent's private view of progress,
open issues, and unresolved risks. It is never shown to the action agent. This
allows the memory agent to maintain a working model of the task without
polluting the action agent's context.

The \textbf{knowledge} memory stores relatively stable facts that are expected
to remain true during the task, such as task requirements, environment
properties, file paths, configuration details, user- or tool-verified facts, and
evaluation-relevant observations. These entries capture what the agent has
learned about the task or environment.

The \textbf{procedural} memory stores attempts and outcomes, such as commands
that failed, fixes that succeeded, hypotheses that were ruled out, diagnostic
signals, or empirical improvements. These entries capture what the agent has
tried and what happened, so that later decisions can avoid repeated failures or
reuse successful evidence.

Each memory entry consists of a short identifier, natural-language content, and
metadata such as creation time and access statistics. Identifiers allow the
memory agent to update or delete stale entries explicitly. In practice, entries
are written in a compact tagged format, such as environment facts, paths, task
facts, bugs, and performance observations. This structure encourages the memory
agent to distinguish constraints from procedural evidence.

\subsection{Two-Phase Memory Agent}

At each memory step, the memory agent runs two phases: memory bank management
and proactive intervention.

\paragraph{Phase 1: Memory management.}
In the first phase, the memory agent manages the memory bank by returning a list
of predefined tool calls. It does not directly rewrite the memory bank. Given the
task description, the recent trajectory window, and the current bank, the memory
agent may call memory\_update\_status, memory\_save\_knowledge,
memory\_save\_procedural, and memory\_delete zero or more times. The system
executes the returned calls in order and treats the resulting bank as the updated
memory bank. If the memory agent returns no tool call, the bank is left
unchanged.

The four calls correspond to the three memory components and deletion.
memory\_update\_status updates the memory agent's private progress tracking,
which is used only by the memory agent and is not shown to the action agent.
memory\_save\_knowledge saves important facts to the knowledge bank, including
task requirements, environment facts, file paths, API details, and key
constraints from the task description. memory\_save\_procedural records
debugging experience, failed approaches, solutions, error patterns, successful
fixes, and performance observations. memory\_delete removes an outdated or
incorrect memory entry by its identifier.

This tool-call interface makes memory management explicit and constrained: the
Phase~1 output is a sequence of bank edits, not a free-form summary of the
trajectory. It lets the memory agent externalize execution state such as
unsatisfied requirements, verified environment facts, failed commands, and
successful fixes while keeping the bank structured across long trajectories.

\paragraph{Phase 2: Intervention selection and transient injection.}
Conditioned on the updated memory bank and the recent trajectory, the memory
agent selects an intervention action for the next action-agent decision. This
phase does not modify the memory bank. Instead, it decides whether any retained
execution state should be reactivated, and if so, how it should be expressed as
a targeted reminder. Formally, the intervention action is either a reminder
$r_t$ or a null intervention $\varnothing$.

If the memory agent emits a reminder, the system supplies $r_t$ to the next
action-agent call as a separate transient memory context. The action agent's
base instructions, tools, and decoding procedure are unchanged; the only change
is the optional memory context provided at call time. If the memory agent selects
the null intervention, the action-agent context is left unchanged.

The null intervention is an explicit action. The memory agent is encouraged to
intervene only when a remembered item is likely to affect the next action-agent
decision. Useful interventions include reminders of requirements that are about
to be violated, environment facts that explain the current observation, previous
attempts that should not be repeated, diagnoses that remain relevant, or open
subgoals that are being neglected. The memory agent is discouraged from giving
broad strategic advice, restating information already visible in the current
observation, or taking over the action agent's planning. Thus, intervention
timing is part of the memory policy rather than a consequence of every memory
update.

\subsection{Triggering the Memory Agent}

The memory agent is invoked according to a trigger function $g(t)$. In our main
implementation, it runs at the first step and then at a fixed interval. This
simple schedule lets the memory agent track the trajectory closely and decide
whether to remain silent or intervene. More selective triggers are possible, such
as invoking memory only after tool errors, failed tests, repeated commands, or
large context shifts, but we use a fixed interval to isolate the effect of the
memory intervention policy itself.

\subsection{Learning Memory Intervention Policies}
The memory-intervention architecture does not require training a new model. In
our main instantiation, the memory agent can be implemented as a prompted model
that follows the two-phase interface described above. However, prompted memory
agents are not necessarily the most practical deployment setting: they add
additional inference cost, and their intervention decisions may be imperfectly
calibrated. We therefore treat training an open-weight memory agent as an early
exploration of whether the same intervention policy can be learned rather than
only prompted.

We train only the memory agent while keeping the action agent fixed. Supervised
fine-tuning distills trajectories from a prompted memory agent, teaching the
model to perform memory-bank operations and to choose between a targeted reminder
and a null intervention. This stage mainly teaches the interface and basic
discipline of memory management: compact writing, updating stale state, and
avoiding unnecessary reminders.

Because imitation alone does not optimize the downstream effect of memory
interventions, we further calibrate the memory agent with reinforcement learning.
The goal is not to maximize memory usage, but to improve the intervention policy:
the trained model should learn when remembered execution state is likely to help
the next action decision, and when silence is preferable. We view this training
study as preliminary evidence that memory intervention can be distilled into an
open-weight policy; the architecture itself remains independent of this training
procedure.
% \begin{comment}
% The key distinction between our approach and ordinary summarization is that the
% memory agent does not merely compress the trajectory. A summarizer asks what
% information should be retained. Our memory agent asks a stronger control
% question: given the current state of the task, should any remembered execution
% state become active in the action agent's next decision?

% This distinction matters because long-horizon tasks vary substantially in their
% failure modes. In some tasks, the useful memory is a hard requirement from the
% initial instruction; in others, it is an environment fact, a previous failed
% command, a bug diagnosis, or an unfinished subgoal. A fixed summarization policy
% cannot know whether a memory should interrupt the next action. Our method learns
% this intervention policy explicitly.
% \end{comment}

\section{Experiments and Results}

\subsection{Benchmarks and Setup}

We evaluate our memory agent on two long-horizon agent benchmarks:
Terminal-Bench 2.0 and $\tau^2$-Bench. Terminal-Bench 2.0 evaluates autonomous agents
operating in realistic command-line environments, where agents must inspect
files, run commands, edit code, debug failures, and satisfy hidden verifier tests
\citep{merrill2026terminalbench}. In this setting, execution state decays through
local debugging loops: agents may forget earlier requirements, repeat failed
commands, overlook environment observations, or lose track of diagnosed errors.

$\tau^2$-Bench evaluates conversational tool-use agents in dynamic task
environments derived from real-world service domains \citep{yao2024taubench,
barres2025tau2bench}. Unlike purely autonomous terminal tasks, these interactions
require the agent to maintain policy-relevant state across a multi-turn
conversation, coordinate with a user simulator, and use tools to update or verify
the environment. This creates a different source of behavioral state decay:
important facts may be stated by the user, observed through tools, or implied by
domain policies, but later stop influencing the agent's decisions.

Together, these benchmarks allow us to test whether memory  helps
across both autonomous execution and interactive coordination. Terminal-Bench
emphasizes environment grounding, debugging continuity, and procedural memory;
$\tau^2$-Bench emphasizes policy adherence, user-state tracking, and
conversation-level execution state.

\paragraph{Terminal-Bench 2.0.}
We run the Terminal-Bench 2.0 full set using the official Terminus-2-style
terminal agent harness. The benchmark contains 89 tasks; we report results on
the 85 paired tasks for which both baseline and memory runs produced valid
evaluations, excluding four docker failures unrelated to agent behavior.
For each task, the action agent interacts with a containerized terminal
environment and receives a binary pass/fail score from the task verifier. We
report pass@1 averaged across tasks. For Terminal-Bench, pass@1 is the fraction of tasks whose final verifier
passes.

\paragraph{$\tau^2$-Bench.}
We evaluate three domains from the $\tau^2$-Bench framework: airline, retail, and
telecom. We use the base task split, containing 50 airline tasks, 114 retail
tasks, and 114 telecom tasks, for a total of 278 tasks per configuration. Each
episode is a single sampled conversation between the action agent and a user
simulator, with task success determined by the benchmark evaluator. We report
pass@1 for each domain and a task-weighted average across domains.
For $\tau^2$-Bench, a conversation is considered a pass if it satisfies the task evaluator.

\begin{table*}[t]
\centering
\small
\setlength{\tabcolsep}{6pt}
\caption{
Main results on Terminal-Bench and $\tau^2$-Bench. All scores are pass@1
percentages; $\Delta$ is reported in percentage points (pp). The memory agent
is Claude Opus 4.6.
}
\label{tab:main_results}
\begin{tabular}{lllrrrr}
\toprule
Benchmark & Domain / Split & Action Model & $n$
& Baseline & + Memory & $\Delta$ \\
\midrule
Terminal-Bench 2.0 & full set & Sonnet 4.5 & 85
& 37.6\% & \textbf{45.9\%} & \textbf{+8.3 pp} \\
Terminal-Bench 2.0 & full set & Opus 4.6 & 85
& 43.5\% & \textbf{45.9\%} & \textbf{+2.4 pp} \\
\midrule
$\tau^2$-Bench & airline & Sonnet 4.5 & 50
& 68.0\% & \textbf{78.0\%} & \textbf{+10.0 pp} \\
$\tau^2$-Bench & retail & Sonnet 4.5 & 114
& 49.1\% & \textbf{58.8\%} & \textbf{+9.6 pp} \\
$\tau^2$-Bench & telecom & Sonnet 4.5 & 114
& 55.3\% & \textbf{57.9\%} & \textbf{+2.6 pp} \\
\cmidrule(lr){2-7}
$\tau^2$-Bench & task-weighted avg. & Sonnet 4.5 & 278
& 55.0\% & \textbf{61.8\%} & \textbf{+6.8 pp} \\
\midrule
$\tau^2$-Bench & airline & Opus 4.6 & 50
& 76.0\% & 76.0\% & +0.0 pp \\
$\tau^2$-Bench & retail & Opus 4.6 & 114
& 64.9\% & \textbf{69.3\%} & \textbf{+4.4 pp} \\
$\tau^2$-Bench & telecom & Opus 4.6 & 114
& 63.2\% & \textbf{64.9\%} & \textbf{+1.8 pp} \\
\cmidrule(lr){2-7}
$\tau^2$-Bench & task-weighted avg. & Opus 4.6 & 278
& 66.2\% & \textbf{68.7\%} & \textbf{+2.5 pp} \\
\bottomrule
\end{tabular}
\end{table*}
\paragraph{Agents and memory configuration.}
We evaluate two action-agent strengths: Claude Sonnet 4.5 and Claude Opus 4.6.
Unless otherwise stated, the memory agent is Claude Opus 4.6. The action agent is
kept unchanged across baseline and memory-enabled runs. In the memory condition,
the memory agent runs at the first step and then at every subsequent step. It observes the task description, a recent trajectory
window of $k$ ($k$=8) messages, and the current memory bank. It first updates the
memory bank, then emits either a context injection or
no intervention. If a reminder is emitted, it is injected into the next action-agent call as transient memory context.

\subsection{Main Results}
Table~\ref{tab:main_results} reports the main results. Memory intervention improves pass@1 on both benchmarks and across both action-agent strengths: on Terminal-Bench 2.0, Sonnet 4.5 gains +8.3 pp ($37.6\% \to 45.9\%$) and Opus 4.6 gains +2.4 pp ($43.5\% \to 45.9\%$); on $\tau^2$-Bench, Sonnet 4.5 gains +6.8 pp on the task-weighted average ($55.0\% \to 61.8\%$) and Opus 4.6 gains +2.5 pp ($66.2\% \to 68.7\%$). Gains are larger for the weaker action agent but do not disappear for the stronger one, indicating that the benefit is not merely compensating for limited capacity. Across $\tau^2$-Bench domains, airline and retail show the largest Sonnet lifts (+10.0 and +9.7 pp) while telecom is smaller (+2.6 pp), consistent with our framing of memory as a domain-sensitive intervention policy rather than a fixed summarization mechanism.

\begin{table*}[t]
\centering
\small
\setlength{\tabcolsep}{4pt}
\caption{
Ablations on $\tau^2$-Bench with Sonnet 4.5 as the action agent, Opus 4.6 as the memory agent for all experiments. Scores are
pass@1 percentages. Macro averages domains equally, micro is task-weighted.
}
\label{tab:ablation_all}
\resizebox{\textwidth}{!}{%
\begin{tabular}{lllccccc}
\toprule
 & & & \multicolumn{3}{c}{$\tau^2$-Bench pass@1 (\%)} & \multicolumn{2}{c}{Average} \\
\cmidrule(lr){4-6} \cmidrule(lr){7-8}
Variant & Phase 1 (memory) & Phase 2 (intervention) & Airline & Retail & Telecom & Macro & Micro \\
\midrule
Sonnet 4.5 baseline
& --- & ---
& 68.0 & 49.1 & 55.3 & 57.5 & 55.0 \\
\midrule
\textbf{+ Full memory agent (ours)}
& bank management
& selective reminder / silence
& \textbf{\imp{78.0}} & \imp{57.0} & \imp{57.9}
& \textbf{\imp{64.3}} & \imp{61.2} \\
+ Full-bank context
& bank management
& expose full bank
& \imp{74.0} & \imp{52.6} & \imp{57.9}
& \imp{61.5} & \imp{58.6} \\
+ Always inject
& bank management
& forced reminder every step
& \imp{72.0} & \imp{58.8} & \imp{59.6}
& \imp{63.5} & \textbf{\imp{61.5}} \\
+ Injection-only (no bank)
& skipped
& selective guidance / silence
& \reg{62.0} & \imp{54.4} & \textbf{\imp{66.7}}
& \imp{61.0} & \imp{60.8} \\
+ Mem0 \citep{mem0ai2026mem0}
& Mem0 ADD
& vector+BM25 top-10
& 68.0 & \textbf{\imp{59.6}} & \imp{58.8}
& \imp{62.1} & \imp{60.8} \\
\bottomrule
\end{tabular}%
}
\end{table*}

\subsection{Ablations and Comparisons}

We ablate the memory agent on $\tau^2$-Bench to test which parts of the architecture are responsible for the gains. Our memory design has two agentic components. Phase~1 performs memory management: it maintains a persistent memory bank of user facts, tool observations, procedural evidence, and open subgoals. Phase~2 performs intervention: it decides whether any remembered state should be reactivated in the next action-agent call, and if so, generates a concise memory-grounded reminder. The ablations remove one capability at a time, allowing us to compare our design against passive context augmentation, always-on memory injection, advisor-style guidance.

\paragraph{Ablation variants.}
\textit{Full-bank context} keeps Phase~1 but removes Phase~2 intervention
selection. The memory agent still performs agentic memory management, but the
entire memory bank is exposed to the action agent at every step. This resembles
passive context augmentation: the system maintains memory, but does not decide
which memory item is currently relevant or whether it should intervene in the
control loop. This tests whether the benefit comes merely from making historical
state visible.

\textit{Always inject} keeps both the memory bank and the generative Phase~2
reminder, but removes the silence action. Unlike embedding-based
retrieval-augmented prompting, which retrieves fixed records from a store
\citep{lewis2020rag}, this variant still uses the memory model to synthesize a
reminder from the maintained bank. However, it must inject at every step. This
tests whether the no-op decision is only an efficiency optimization or an
essential part of intervention calibration.

\textit{Injection-only (no bank)} removes persistent memory while retaining the
Phase~2 intervention model. This is analogous to advisor-style systems, where an
auxiliary model observes the trajectory and provides natural-language guidance
to an executor \citep{asawa2025advisor,anthropic2026advisor}. Unlike the full
memory agent, this variant has no maintained memory bank, so its guidance cannot
be grounded in persistent execution state.

\textit{Mem0} compares against a general persistent memory layer
\citep{mem0ai2026mem0}. In this variant, memories are written through Mem0's
ADD interface and retrieved with its vector/BM25 search, with the top retrieved
items returned to the action agent. This isolates the difference between
production memory retrieval and our formulation of memory as a selective
intervention policy.

Table~\ref{tab:ablation_all} shows that memory-style variants generally improve
over the Sonnet baseline, but the improvements are not equally robust across
domains. The full two-phase memory agent achieves the highest macro-average,
improving all three domains and giving the largest gain on airline. Since macro
averages domains equally, this indicates that our architecture gives the most
balanced improvement across heterogeneous task types.

The ablations show that each removed capability weakens robustness in a
different way. Full-bank context improves over the baseline but trails the full
system by 2.8 macro points and 2.6 micro points, suggesting that maintaining a
memory bank is useful but exposing all remembered state is not enough; the memory
agent must select what is currently behaviorally relevant. Always inject is
competitive and slightly leads on micro-average by 0.3 points, but this gap is
within expected run variance and disappears on the domain-balanced macro-average,
where selective silence is better, especially on airline. Injection-only guidance helps in some domains, especially
telecom, but is less stable without persistent memory and even hurts airline
relative to the baseline. Together, these results suggest that neither passive
memory exposure, always-on reminders, nor generic auxiliary guidance is
sufficient: the most balanced gains come from combining maintained execution-state
memory with a selective intervention policy.

Mem0 improves the average score, confirming that persistent
memory retrieval is useful. However, it does not improve airline over the
baseline and falls short of our full system on macro-average. This highlights
the distinction between retrieving memories and deciding whether retrieved state
should intervene. A general memory layer can return relevant records, but it
does not explicitly model when a memory should enter the loop or how it
should be phrased as a targeted reminder.

Overall, the ablations support the two-phase design. Agentic memory management
alone is helpful but insufficient; advisor-like intervention without memory is
not reliably grounded; and always-on injection can be competitive but less
balanced. The full memory agent performs best on macro-average because it
combines maintained execution state with a selective intervention policy that can
choose both what to say and when to remain silent.

\subsection{Qualitative Analysis}

We qualitatively inspect memory-enabled trajectories to understand when memory
injection changes agent behavior. Across both benchmarks, the dominant pattern
is not generic summarization. Memory helps when it reactivates a specific piece
of execution state at the moment the action agent is about to ignore it. The
type of state differs by benchmark: Terminal-Bench emphasizes environment
observations, debugging continuity, and failure diagnoses, while $\tau^2$-Bench
emphasizes policy clauses, user/tool state, and procedural sequencing.
Table~\ref{tab:qual_mechanisms} summarizes the recurring mechanisms we observe;
the rest of this section discusses each benchmark in turn.

\begin{table*}[t]
\centering
\small
\setlength{\tabcolsep}{5pt}
\caption{
Representative mechanisms observed in qualitative analysis. Memory interventions
help when they reactivate execution state that the action agent previously saw
but no longer uses in its next decision.
}
\label{tab:qual_mechanisms}
\begin{tabularx}{\textwidth}{l X X}
\toprule
Mechanism & Typical memory content & Representative examples \\
\midrule
Requirement / policy reactivation
& Task or domain rule about an allowed action
& $\tau^2$ airline compensation, retail modification rules \\
Environment grounding
& Runtime facts, paths, tool limitations, system quirks
& Terminal-Bench Git server setup, ARS file-write failure \\
Failure-loop avoidance
& Previous attempts and why they failed
& Adaptive rejection sampling, telecom diagnostic retries \\
Diagnostic carryover
& Root cause of a bug or negative signal
& Regex edge cases, SQLite gcov configuration \\
Progress / entity tracking
& Which user, order, line, branch, or subgoal is active
& $\tau^2$ telecom line lookup, retail authentication state \\
\bottomrule
\end{tabularx}
\end{table*}

\paragraph{Terminal-Bench: maintaining debugging continuity.}
In Terminal-Bench, memory is most useful during iterative debugging, where the
action agent accumulates observations but later loses track of which ones should
constrain the next command. The relevant memory is not a single type of
fact. In \texttt{regex-log}, memory reactivates both requirements and diagnoses:
it points out that the current regex violates the task's boundary condition,
misses single-digit IPv4 octets. In \texttt{adaptive-rejection-sampler}, memory
tracks repeated failed file edits and later surfaces an environment-specific
workaround. These cases suggest that memory improves Terminal-Bench performance
by preserving continuity across debugging iterations: what was tried, what was
observed, what failed, and what still constrains the solution.

\paragraph{$\tau^2$-Bench: reactivating policy and interaction state.}
In $\tau^2$-Bench, the useful execution state is often policy- or
interaction-specific. Airline and retail tasks frequently require the agent to
apply a rule many turns after the relevant fact was observed. For example, in
one airline case, the user claimed Gold status, but tool output showed that the
user was a Regular member. The baseline issued compensation based on the user's
claim, while the memory-enabled agent reminded the action agent to rely on
verified records. In another case, memory prevented an invalid modification of a
basic-economy flight by reactivating the policy clause that such flights cannot
be modified. Successful interventions often occur immediately before a
state-changing tool call, reminding the agent of authentication requirements,
eligibility conditions, one-shot tool limits, or policy clauses not enforced.

Overall, this qualitative analysis aligns with the main results: memory
interventions help more often than they hurt. Successful interventions are
specific, grounded, and timely; they restore the causal influence of execution
state over the next action. The remaining failures are primarily calibration
errors rather than failures of memory storage. Occasionally, the memory agent
surfaces a speculative inference with too much confidence, repeats information
that the action agent already knows, or raises a plausible but unnecessary
concern that causes extra verification. These cases motivate training the memory
agent to better decide when to remain silent.

\subsection{Training Open-Weight Memory Agents}
\label{sec:training}

We conduct an early exploration of whether the memory-intervention policy can be
trained on an open-weight model. The prompted Opus memory agent improves
long-horizon agents, but it adds a frontier-model call at each memory step and
its intervention calibration is imperfect. We therefore train Qwen3.5-27B as a
memory agent while keeping the Qwen3.5-122B-A10B action agent frozen.

We train on SETA, a collection of executable terminal-agent tasks with verifier
rewards \citep{seta}. SETA is used for supervised fine-tuning and reinforcement
learning, and Terminal-Bench 2.0 is held out to test transfer. SFT distills
prompted-memory trajectories, supervising both Phase~1 memory-bank operations
and Phase~2 intervention decisions. GRPO then further optimizes intervention
calibration. Because task-level verifier rewards are sparse over many memory
calls, we focus updates on pivot turns that labeled offline rollouts identify as likely to affect downstream success.

\begin{table}[t]
\centering
\small
\caption{
Training and transfer of the open-weight memory agent. The action agent is a
frozen Qwen3.5-122B-A10B; the trainable memory agent is Qwen3.5-27B. SETA
validation reports average verifier reward; Terminal-Bench reports pass@1 on
the 85-task.
}
\label{tab:trained_memory}

\begin{subtable}[t]{0.49\textwidth}
\centering
\setlength{\tabcolsep}{5pt}
\caption{SETA validation.}
\label{tab:seta_trained_memory}
\begin{adjustbox}{max width=\linewidth}
\begin{tabular}{lccc}
\toprule
Setup & Avg. reward & Solved & $\Delta$ \\
\midrule
Action only, no memory
& 0.709 & 56 & --- \\
+ Qwen3.5-27B base memory
& 0.693 & 54 & -0.016 \\
+ SFT memory
& 0.720 & 58 & +0.011 \\
+ GRPO memory
& \textbf{0.734} & \textbf{58} & \textbf{+0.025} \\
\bottomrule
\end{tabular}
\end{adjustbox}
\end{subtable}
\hfill
\begin{subtable}[t]{0.49\textwidth}
\centering
\setlength{\tabcolsep}{6pt}
\caption{Transfer to Terminal-Bench 2.0.}
\label{tab:trained_memory_terminalbench}
\begin{adjustbox}{max width=\linewidth}
\begin{tabular}{lccc}
\toprule
Setup & $n$ & Pass@1 & $\Delta$ \\
\midrule
Qwen3.5-122B-A10B action only
& 85 & 37.6\% & --- \\
+ trained Qwen3.5-27B memory
& 85 & \textbf{41.1\%} & \textbf{+3.5 pp} \\
\bottomrule
\end{tabular}
\end{adjustbox}
\end{subtable}
\end{table}

Table~\ref{tab:seta_trained_memory} shows that the memory policy is learnable
but requires calibration. An untrained 27B memory agent hurts performance,
reducing SETA reward from 0.709 to 0.693. SFT recovers this loss and improves
reward to 0.720, while GRPO further raises it to 0.734. On held-out
Terminal-Bench 2.0, the SETA-trained memory agent improves the frozen action
agent from 37.6\% to 41.1\% pass@1. These results suggest that memory
intervention can be distilled into an open-weight model and that RL improves the
decision of when remembered state should enter the control loop.
\section{Conclusion}
\label{section:conclusion}

We propose agent memory as proactive intervention policy rather than passive storage and retrieval. Long-horizon agents suffer from \emph{behavioral state decay}, where execution state that should guide future actions stops influencing behavior. Our memory agent maintains this state and injects grounded context when they are likely to affect the next decision. Across Terminal-Bench 2.0 and $\tau^2$-Bench, our memory agent consistently improves task success across different action agent models. Training on SETA further shows that this intervention policy is learnable: SFT teaches memory management, while RL improves when remembered state should re-enter the control loop. Open directions include jointly training the memory and action agents, learning when memory is invoked rather than running on a fixed schedule, and identifying when verbatim reminders versus task-specific abstractions are most effective.
\bibliographystyle{assets/plainnat}
\bibliography{paper}

%\clearpage
%\newpage
%\beginappendix
%\input{sections/8_appendix}

\end{document}